\documentclass[utf8]{article} % for Science, Engineering and Humanities and Social Sciences articles
\usepackage{url,hyperref,lineno,microtype,subcaption}
\usepackage[onehalfspacing]{setspace}

\usepackage[affil-it,blocks]{authblk}
\usepackage[numbers]{natbib}

\usepackage[colorinlistoftodos,disable]{todonotes}

\graphicspath{{figs/}}
\usepackage{pdflscape}
\usepackage{booktabs}
\usepackage{tabulary}
\usepackage{soul}
\usepackage{color}

\usepackage{xspace}
\newcommand{\method}{Led-by-Experts Automation and Design Of Robots\xspace}
\newcommand{\meth}{LEADOR\xspace}

\title{LEADOR: A Method for End-to-End Participatory Design of Autonomous Social Robots}

\author{Katie Winkle}
\affil{KTH Royal Institute of Technology, Sweden}

\author{Emmanuel Senft}
\affil{University of Wisconsin-Madison, USA}

\author{S\'{e}verin Lemaignan}
\affil{Bristol Robotics Laboratory, UK}

\date{\today}

\newcommand\blfootnote[1]{%
  \begingroup
  \renewcommand\thefootnote{}\footnote{#1}%
  \addtocounter{footnote}{-1}%
  \endgroup
}

\begin{document}

%\maketitle
\makeatletter
\begin{titlepage}%
  \let\footnotesize\small
  \let\footnoterule\relax
  \blfootnote{Katie Winkle and Emmanuel Senft contributed equally to this paper and share first authorship. Correspondence to winkle@kth.se and esenft@wisc.edu respectively.}
  \let \footnote \thanks
  \null\vfil
  \vskip 60\p@
  \begin{center}%
    {\LARGE \@title \par}%
    \vskip 3em%
    {\large
     \lineskip .75em%
      \begin{tabular}[t]{c}%
        \@author
      \end{tabular}\par}%
      \vskip 1.5em%
    {\large \@date \par}%       % Set date in \large size.
  \end{center}\par
  \@thanks
  \vfil\null
\end{titlepage}%
\makeatother

\onecolumn
% \title{LEADOR: A Method for End-to-End Participatory Design of Autonomous Social Robots} 

% \maketitle

\begin{abstract}
Participatory Design has been used to good success in Human-Robot Interaction (HRI),
but typically remains limited to the early phases of development, with
subsequent robot behaviours then being hardcoded by engineers or utilised in
Wizard-of-Oz (WoZ) systems that rarely achieve autonomy. In this article, we
present \meth (\method) an \textit{end-to-end} participatory design (PD) methodology for domain
expert co-design, automation and evaluation of social robot behaviour. This
method starts with typical PD, working with the
domain expert(s) to co-design the interaction specifications and state and action
space of the robot%, and an interface for their interaction with it. 
It then replaces the traditional offline programming or WoZ phase by an in-situ and
online teaching phase where the domain expert can live-program or
teach the robot how to behave while being embedded in the interaction
context. We believe that this live teaching phase can be best achieved
by adding a learning component to a WoZ setup, which captures experts'
implicit knowledge, as they intuitively respond to the dynamics of the
situation. The robot then progressively learns an appropriate, expert-approved
policy, ultimately leading to full autonomy, even in sensitive and/or
ill-defined environments. However, \meth is agnostic to the exact technical approach used to facilitate this learning process. The extensive inclusion of the domain expert(s) in
robot design represents established responsible innovation practice, lending
credibility to the system both during the teaching phase and when operating
autonomously. The combination of this expert inclusion with the focus on in-situ development also means \meth supports a mutual shaping approach to social robotics.  %We draw on the two previously published case studies from which this method has been derived (an
% educational robot for school children and a fitness coach robot for adults)
% to offer a blueprint for applying this methodology in practice, as well as
% to identify limitations and opportunities when applying this framework in
% new environments.
We draw on two previously published, foundational works from which this
(generalisable) methodology has been derived in order to demonstrate the
feasibility and worth of this approach, provide concrete examples in its
application and identify limitations and opportunities when applying this
framework in new environments.

\end{abstract}

\section{Introduction}

In the context of robotics research, participatory design attempts to empower
non-roboticists such that they can shape the direction of robotics research and
actively collaborate in robot design~\citep{leeStepsParticipatoryDesign2017}.
Typically, participatory design is achieved by researchers running workshops or focus groups
with end-users and/or domain experts. Output may include potential use case
scenarios~\citep{jenkinsCareMonitoringCompanionship2015}, design guidelines/
recommendations~\citep{winkleSocialRobotsEngagement2018} and/or prototype robot
behaviours~\citep{azenkotEnablingBuildingService2016}. {\v S}abanovi{\'c} identified such
methods as appropriate for the pursuit of a mutual shaping approach in robot
design, that is one which recognises the dynamic interactions between social
robots and their context of use~\citep{sabanovicRobotsSocietySociety2010}, an
approach that we find compelling for designing effective and acceptable social
robots efficiently. However, the automation of social robot behaviour, which
requires a significant technical understanding of robotics and artificial
intelligence, is not typically considered during such activities. 

Instead, common methods for the automation of social robot behaviour include
utilising models based on human psychology (e.g. Theory of
Mind~\citep{lemaignanArtificialCognitionSocial2017}) or animal
behaviour~\citep{arkinEthologicalModelingArchitecture2001}, or attempting to
observe and replicate human-human interaction behaviours
(e.g.~\citep{sussenbachRobotFitnessCompanion2014}). %, or more simply designing a behaviour based on a set of heuristics. 
This limits the potential for
direct input from domain experts (teachers, therapists etc.) who are skilled in
the use of social interaction in complex scenarios. Previous work with such experts has demonstrated that a
lot of the related expertise is intuitive and intangible, making it difficult to
access in a way that can easily inform robot automation~\citep{winkleSocialRobotsEngagement2018}. This is somewhat
addressed by methods that capture domain expert operation of a robot directly,
for example end-user programming tools (e.g.~\citet{leonardi2019trigger}) or
learning from expert teleoperation of robots
(e.g.~\citet{sequeira2016discovering}). However, these methods
tend to focus on offline learning/programming. As such, there is no opportunity
for experts to create an adequate, situated mental model of the robot's
capabilities, limiting the guarantee of appropriate behaviour when the robot is eventually
deployed to interact with users autonomously. 

Instead, we argue that robots should be automated by domain experts
themselves, in real-time, and while being situated in the interaction context; and
that this automation should be done through a direct, bi-directional interaction between the expert
and the robot. We refer to this as the \emph{teaching phase}, where the robot is
taught what to do by the domain expert, regardless of whether it is, e.g. a
machine learning algorithm or an authoring tool that underpins this interaction. This live,
in-situ and interactive approach allows \emph{mutual shaping} to occur during robot
automation, as the expert defines the robot's program in response to the
evolving dynamics of the social context into which the robot has been deployed.
%between the robot and the expert defining their program, as the expert  %which delivers on {\v S}abanovi{\'c} point of ??
%interaction between the domain expert `teacher' and the robot `learner', meaning the only way to evaluate learned behaviours is to test them. In addition, such methods are not often typically integrated into a fully participatory design process, which would also, for example, involve the domain-experts in the design of those tools designed to capture their expertise.   

\smallskip

\subsection{Supporting A Mutual Shaping Approach to Robot Design}

{\v S}abanovi{\'c} proposed a \emph{mutual shaping} approach to social robot design, that is one
which recognises the dynamic interactions between social robots and their
context of use, in response to their finding that most roboticists were taking a
technologically deterministic view of the interaction between robots and
society~\citep{sabanovicRobotsSocietySociety2010}. Studies of real-world HRI
motivate such an approach, as they demonstrate how mutual shaping effects impact robot effectiveness upon deployment in the real world. For example, use and
acceptance of robots in older adult health settings has been shown to be
affected by situational and context of use factors such as user age and gender,
household type, and the prompting of its use by
others~\citep{degraafSharingLifeHarvey2015,changInteractionExpandsFunction2015}
i.e. factors unrelated to the robot's functionality. Pursuit of a mutual shaping
approach, primarily through use of participatory design and in-the-wild robot
evaluation methods, gives the best possible chance of identifying and accounting
for such factors during the design and development process, such that the robot
has maximum positive impact on its eventual long-term deployment.  

To this end, {\v S}abanovi{\'c} describes four key practices that underpin a mutual
shaping approach to support a \textit{``socially robust understanding of
technological development that enables the participation of multiple
stakeholders and disciplines''}: 

\begin{enumerate}
    \item Evaluating robots in society: human-robot interaction studies and
        robot evaluations should be conducted `in the wild', i.e. in the
        environments and context of use for which they are ultimately intended
        to be deployed~\citep{ros2011child}.

    \item Studying socio-technological ecologies: robot design should be
        informed by systematic study of the context of use, and evaluation of
        robots should consider impact on the socio-technology ecology into which
        they have been deployed.  

    \item Outside-in design: design constraints should be defined by empirical
        social research and the social context of use, rather than technical
        capabilities, and evaluation should be based on user experiences rather
        than internal measures of technical capability.

    \item Designing with users: stakeholders (those who will be directly
        affected by the robot's deployment and use) should be included in
        identifying robot applications and thinking about how robots will be
        used as well as in designing the robot and its behaviour(s).

\end{enumerate}

% By combining best practices from participatory design with human-in-the-loop
% machine learning processes, it is possible to deliver on one or more of these
% practices at particular stages of robot development. 
However, as we explain in
Section~\ref{sec:related}, robot development at present typically represents a
discontinuous process, particularly broken up by the automation of social
robot behaviour. It still tends to heavily rely on technical expertise, executed in research/development
environments rather than the real-world, with little active inclusion of
domain experts or other expert stakeholders. This discontinuity also represents
a key hurdle to truly multi-disciplinary working, a disconnect between those of
different academic backgrounds on the research team which can result in a number
of practical challenges and frustrations. %as we reflect on in Table
%\ref{tab:typicalPD}.

\subsection{The \emph{\method} (\meth) Method}

The generalisable method we provide in this work derives from two (independently undertaken)
foundational works. Firstly is~\cite{senft2019teaching}'s educational robot for
school children, in which a psychologist taught a robot to support children in an
educational activity. After the teaching phase with 25 children, the robot was evaluated in
further autonomous interaction with children which demonstrated the opportunity
of online teaching as a way to define autonomous robot behaviours.

Secondly
is~\cite{winkle2020insitu}'s robot fitness coach. This work built
on~\cite{senft2019teaching} by integrating the online teaching method into an
end-to-end participatory design process whereby the same professional fitness instructor was
involved in the co-design, automation and evaluation of a robot fitness coach.
%, deployed to deliver a real-world (NHS) exercise programme in a real gym
%environment. 
This work also demonstrated the value of online teaching when compared to
expert-designed heuristics as a next-best alternative for defining autonomous
robot behaviours with domain expert involvement. Both studies used a teaching
phase where a domain expert interacted with the robot to create an interactive
behaviour, and in both studies, the resulting autonomous robot behaviour was
evaluated with success.
%in following interactions where the robot was fully autonomous.

\begin{figure}
  \centering
  \includegraphics[width=\linewidth]{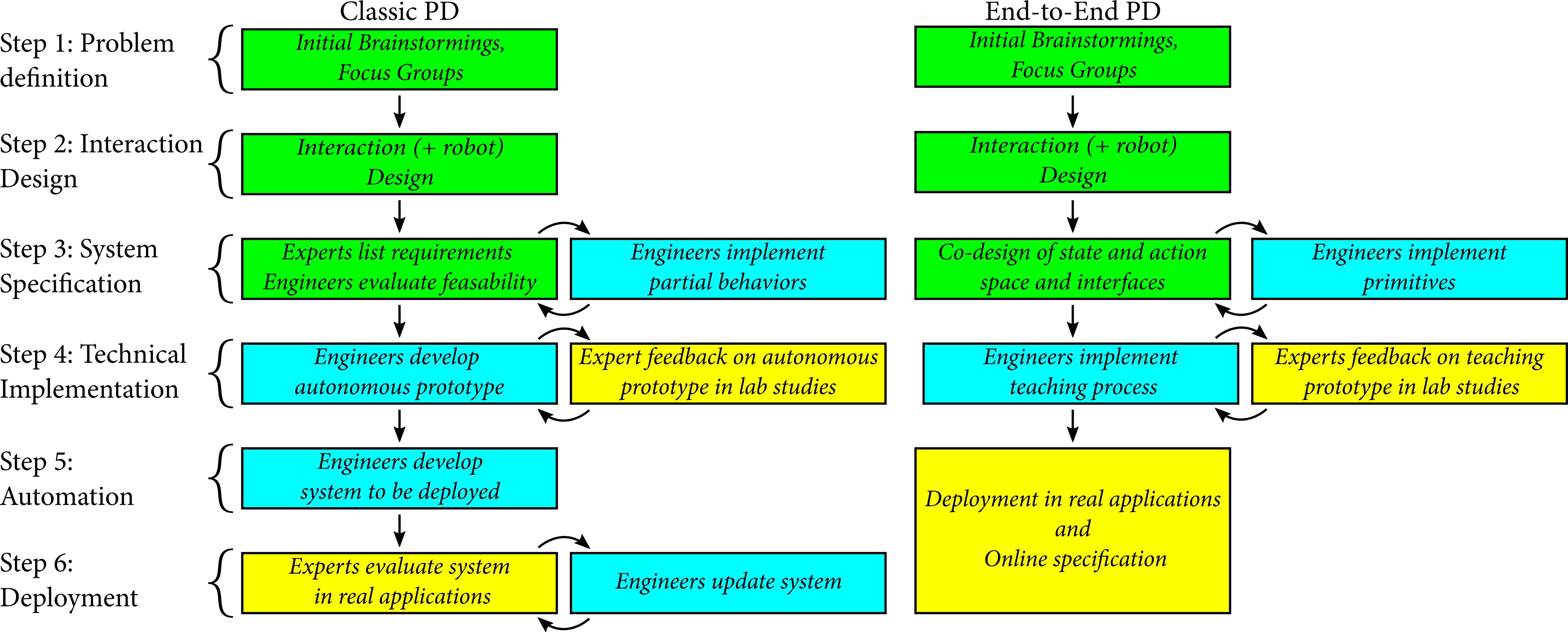}
%   \caption{An end to end methodology for participatory design of autonomous social robots. Note that the use of online teaching for program definition essentially allows steps 5 and 6 to be merged and conducted in parallel. Alternative methods for program definition and user testing might be used, but online teaching specifically suits situations where (i) the (human expert) reasoning for particular action selection/evaluation is highly intuitive, (ii) the context of use requires continuously appropriate/effective robot behaviour. \katie{boxes are long steps, so keep 5 and 6 as they are on the left; the right side is some other chart shapes to represent the instantaneous nature of these feedback, and highlight that we can do 5 and 6 at the same time. Could colour code what is engineer and what is domain expert and what is both - note that right hand side is *only* domain expert + robot. Change the diagram such that it's left side VS right side rather than right side == left side.}}
  \caption{Comparison between a classic Participatory Design (PD) approach and
      \meth, our proposed end-to-end participatory design approach. Green
      activities represent joint work between domain experts, multidisciplinary
      researchers and/or engineers; yellow activities are domain expert-led;
      blue activities are engineer-led. Compared to typical PD, the two key differences in our approach
      are the focus on developing a \emph{teaching system} instead of a
      \emph{final autonomous behaviour} in step 4, and the combining of autonomous action
      policy definition and deployment in the real world into a single step 5 +
      6. In addition, our method reduces the amount of work that is carried out
      independently by engineers (i.e. with no domain expert or non-roboticist
      input).}

  \label{fig:methodcomp}
\end{figure}

From these works, we have derived a five step, generalisable, method for
end-to-end participatory design (PD) of autonomous social robots (\emph{\method}
or \meth), depicted alongside typical PD in Figure~\ref{fig:methodcomp}. The key
stages of our approach, as referenced in the figure, can be summarised as
follows: 

\begin{itemize}
    \item[(i)] Problem Definition:  \textit{Initial brainstorming, studies of context of use, studies with stakeholders.}
    \item[(ii)] Interaction Design: \textit{Detailed refinement of robot application and interaction scenario, choice/design of robot platform.}
    \item[(iii)] System Specification: \textit{Co-design of the robot's action space, input space, and teaching interface.}
    \item[(iv)] Technical Implementation: \textit{Realisation of (iii) through technical implementation of underlying architecture and all sub-components and tools required for the teaching phase.}
    \item[(v)] Real World Deployment: \textit{Robot is deployed in the real-world, where a teaching phase is undertaken, led by the domain expert(s), to create autonomous robot behaviour.}
\end{itemize}

% Notably, compared to typical PD, the key contribution of our method is that the expert-led teaching phase during real world deployment results in (i) automation occuring in the real world, live and online
% in response to the social specifities of the context of use, and (ii) the
% robot automation process becoming a two-way interaction between the expert and
% the robot system, which itself takes place within three-way interactions between
% the expert, the robot, and the target user as
% well as the broader interactions between the deployed system and its
% environment (see Figure \ref{fig:three-way}).

\begin{figure}
  \centering
  \includegraphics[width=.5\linewidth]{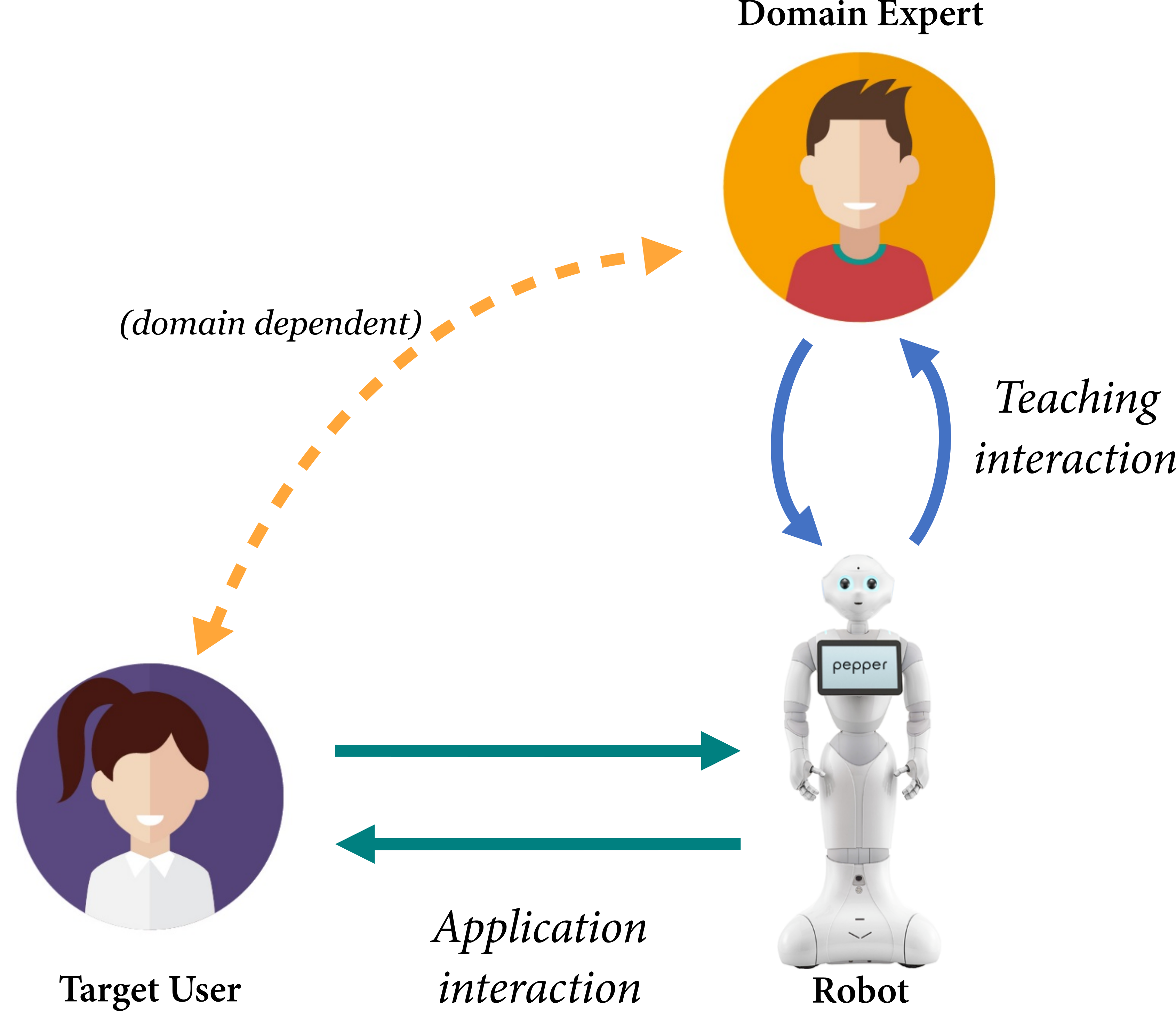}
 \caption{Three-way interaction between the domain-expert, the robot, and the
 target user through which the expert teaches the robot during a teaching phase
 upon real world deployment. Robot automation is therefore happening in the real
 world, while the robot is fully embedded in its long-term application context.
 The expert is teaching the robot through bi-directional communication, as the
 robot interacts with the target user. The extent of interaction(s) between the
 domain expert and target user should be consistent with what is envisaged for
 long term deployment of the robot, and is domain-dependent.} %, as per Figure~\ref{fig:autonomy-exper}.}

  \label{fig:three-way}
\end{figure}

The cornerstone of our method is to facilitate robot automation through
direct interaction between the expert and the robot, during a `teaching phase'
whereby the domain expert teaches the robot what to do during real interaction(s) with the target user. The resultant
interaction is depicted in Figure~\ref{fig:three-way}.
Regardless of the specifics of the final interaction, the output of
this phase is a robot that \textit{can} operate autonomously, but could also
allow for continued expert-in-the-loop supervision and/or behaviour
correction/additional training.

Through our foundational works, we demonstrate the flexibility in our method for developing autonomous robots for different long-term interaction settings. \citet{senft2019teaching}'s educational robot  was intended for diadic, unsupervised robot-user interactions, whereas \citet{winkle2020insitu}'s robot fitness coach was intended primarily for diadic robot-user interactions but to be complimented with additional expert-user interactions/supervision and with additional expert-robot-user teaching
interactions if necessary. \meth could also be used to design robots with other interaction requirements, e.g. an autonomous robot to be used in fully triadic expert-robot-user interactions or to facilitate permanent expert supervision and validation of autonomous behaviour.

In this paper, we have combined our experiences from these foundational works to propose an end-to-end participatory design process, centered around an in-situ teaching phase, that
uniquely delivers on the promises of mutual shaping and participatory design. We suggest this approach is as \emph{practical} as it is \emph{responsible}, as our foundational studies demonstrate we were able to create appropriate, intelligent, autonomous social robot behaviour for complex application environments in a timely manner. As detailed in~\citet{senft2015sparc,senft2019teaching} this teaching phase
is achieved by deploying the robot in the proposed use case and
it being initially controlled completely by a human `teacher'. The teacher
can progressively improve the robot behaviour in-situ and generate a mental
model of the robot's policy.
% 's
% behaviour can then be combined with sensory data to generate training data for
% online machine learning. When enough examples have been captured, the robot can
% start making behavioural suggestions, on which the teacher can provide feedback
This teaching can continue until the domain expert is 
confident that the robot can satisfactorily operate autonomously.
This approach therefore allows non-roboticist, domain experts to actively
participate in creating autonomous robot behaviour. It also allows for the
continual shaping of robot behaviour, as teaching can be seamlessly
(re-)continued at any time to address any changes in the interaction dynamics,
therefore better supporting a mutual shaping approach. We suggest our
methodology%, particularly combined with this online, human-in-the-loop approach
%to autonomy, 
is particularly appropriate for use cases in which difficult-to-automate and/or
difficult-to-explain `intuitive' human domain expertise and experience are
needed to inform personalised interaction and engagement (e.g. socially assistive robotics). The result then, is an autonomous robot which has been designed, developed and
evaluated (by a multi-disciplinary research team) directly in conjunction with
domain experts, within its real-world context of use, that can intelligently
respond to complex social dynamics in ways that would have otherwise been very
difficult to automate.   

For clarity, hereafter we use the term \emph{domain expert} (or \emph{teacher})
to refer to experts in an application domain. For example, these domain experts
could be therapists, shop owners, or school teachers. These experts
interact with the robot and specify its behaviour in a \emph{teaching
interaction} (even if no actual machine learning might be involved). On the
other hand, \emph{engineers} or developers refer to people with technical
expertise in robotics or programming. They are the ones typically programming a
robot behaviour or developing tools to be used by domain experts. Finally, the
\emph{target user} is the person a robot would interact with in the
\emph{application interaction}. For example, such target users could be children
during a therapy session or store customers in a shopping interaction. This
population is expected to interact with the robot at the point of use, rather
than be the ones directly defining the autonomous robot behaviour. 

% \subsection{Case Studies}
% We draw on the two previously published case studies from which this (generalisable) methodology has been derived, and use them to (i) demonstrate the feasibility and worth of this approach and (ii) provide concrete examples in its
% application.

\section{Related Work}
\label{sec:related}

\subsection{Participatory Design and Social Robotics}
\label{sec:relatedPD}
To properly situate our method in the context of participatory design (PD), we
first define PD and how it relates to other methodologies typically seen in
social robotics. Most works relevant to PD in  social human-robot interaction showcase one (or
more) of the following methods:

\begin{enumerate}
    % \item \textit{Laboratory Studies} The overwhelming majority of social
    %     robotics literature describes controlled, laboratory or even video-based
    %     experiments in which researchers aim to statistically evaluate the
    %     impact of particular robot design/behaviour choices (e.g. portraying
    %     emotion~\citep{koschateOvercomingUncannyValley2016,ullrichRobotPersonalityInsights2017,
    %     parkRobotBehaviorExpressions2010}) on a range of measures which might be
    %     objective (e.g.
    %     persuasiveness~\citep{winkleEffectivePersuasionStrategies2019}, task
    %     performance~\citep{kuchenbrandtKeepEyeTask2014}), or more subjective
    %     (e.g. the commonly used `Godspeed'
    %     questionnaire~\citep{bartneckMeasurementInstrumentsAnthropomorphism2009}). 

    \item \textit{Ethnographic/`In-the-Wild' Studies} typically focus on
        understanding situated use and/or emergent behaviour(s) on deployment of
        a robot into the real world. Concerning robot design, such studies are
        inherently limited to the testing of prototypes, WoZ systems or finished
        products (e.g.~\citet{forlizziHowRoboticProducts2007,
        changInteractionExpandsFunction2015}). However, they might be used to
        inform initial design requirements (and their iteration) through
        observation of the use case environment and user behaviour.

    \item \textit{User-Centered Design (UCD)} aims to understand and incorporate
        user perspective and needs into robot design. Typically researchers set
        the research agenda based on prior assumptions regarding the context of
        use and proposed robot application (e.g.
        \citet{louieFocusGroupStudy2014,wuDesigningRobotsElderly2012,beerDomesticatedRobotDesign2012}). 

    \item \textit{Participatory Design (PD)} encourages participants (users,
        stakeholders etc.) to actively join in decision making processes which
        shape robot design and/or the direction of research. This typically
        involves participants having equal authority as the researchers and
        designers, with both engaging in a two-way exchange of knowledge and
        ideas (e.g. \citet{azenkotEnablingBuildingService2016,
        bjorlingParticipatoryResearchPrinciples2019}).

\end{enumerate}

Lee et al. give a good overview of the above practices as employed in social
robot and HRI design/research, with a particular focus on how the shortcomings
of 1 and 2 can be addressed using PD \citep{leeStepsParticipatoryDesign2017}. The
authors use a case study of social robot PD from their own work to highlight a number of PD design principles for
informing social robot design and further development of PD methodologies. They
particularly highlight the empowering of PD participants to become active `robot
co-designers' through \textit{mutual learning}, whereby there is a two way
exchange of knowledge and experience between researchers/designers and expert
stakeholders. Through this process, users learn about e.g. robot capabilities,
such that they are better informed to contribute to discussions on potential
applications, whilst the researchers/designers come to learn more about the
realities of the proposed context of use from the users' perspective.  

Since publication of Lee et al.'s work, PD methods have been gaining visibility for
the design of social robots, with other roboticists further refining PD methods
and best practice for their use in social robotics and HRI. As such, PD works
relating to ours can be grouped into two categories: 

\begin{itemize}
    \item[(i)] results-focused publications which utilised PD methods. 

    \item[(ii)] methodology-focused publications in which the authors share or
        reflect on PD methods for use in social robot/HRI design.%, often advocating for the use of such methods more broadly.

    %\item[(iii)] (typically technology-centred) works where the authors have utilised methods that might be considered/could support PD without explicitly describing it as such
\end{itemize}

Works on (i) have typically taken the form of researchers working closely
with prospective users and/or other stakeholders via focus groups, interviews,
workshops etc. with the researchers then concatenating their results to produce
potential use case scenarios~\citep{jenkinsCareMonitoringCompanionship2015},
design guidelines/recommendations~\citep{winkleSocialRobotsEngagement2018}
and/or prototype robot behaviours~\citep{azenkotEnablingBuildingService2016}.
For example, \citet{azenkotEnablingBuildingService2016} used
participatory design to generate specifications for a socially assistive robot
for guiding blind people through a building. The authors' study consisted of
multiple sessions including interviews, a group workshop and individual
user-robot prototyping sessions. The initial interviews were used, in part, to
brief participants about robot capabilities. The group session was used to
develop a conceptual storyboard of robot use, identifying interactions between
the robot guide and the user. 
% The researchers also asked participants to
% instruct a naive human guide, asking probing questions around their preferences
% and instructions. Finally, participants were individually invited to work with a
% researcher to prototype exemplar behaviours on a robot.

\citet{winkleSocialRobotsEngagement2018} conducted a study with therapists, utilising a novel focus group
methodology combined with follow-up individual interviews in order to generate
an expert-informed set of design guidelines for informing the design and use of
socially assistive robots in rehabilitative
therapies. The topic guides for each
part of the study were designed to help the researchers understand typical
therapy practice and therapist best practices for improving patient engagement
and to explore therapists' ideas and opinions on the potential role(s) social
robots might play in rehabilitation. A key finding of this work was the extent
to which therapists' intuitive, instantaneous behaviour is driven by situational
factors specific to each individual client, making it difficult, for example, to
extract any clear cut heuristics that might inform generalisable, autonomous
social robot behaviour directly. The resultant design guidelines therefore
suggested that socially assistive robots require 'high-level' personalisation to
each user as well as the ability to adapt, in real time, to e.g. the user's
performance and other situational factors. This is one of the key works that
motivates our effort to therefore facilitate expert-led, in-situ robot teaching,
in order to capture this sort of tacit social intelligence.  

A follow up publication by the same authors then comes under category (ii).
Specifically, the authors provide more detail on their focus group methodology,
and how it reflects a mutual shaping approach to social robot design, alongside
a general guide in how it might be applied to other
domains~\citep{winkleMutualShapingDesign2019}. The method combines elements of
PD and UCD, and utilises a demonstration of robot capabilities to support mutual
learning between the researchers and participants. To evidence how this method
supported mutual shaping in their work, and why this was beneficial, the authors identify specific
project-related considerations as well as new research directions that could
only be identified in conjunction with their domain expert participants, and
also note that taking part in a focus group significantly (positively) impacted
on participants' acceptance of social robots.   

Further on (ii), \citet{bjorlingParticipatoryResearchPrinciples2019} shared PD methods they used in the context of
taking an overall \textit{human-centered design} approach to co-designing robots
for improved mental health with
teens. They present three
method cases which cover novel and creative participatory techniques such as
research design, script-writing and prototyping, concluding with a set of
participatory design principles for guiding design work with vulnerable
participants in a human-centred domain. One of their methods revolved around
inviting teens to act as WoZ robot operators. Specifically, their setup had one
teen teleoperating a robot whilst another teen recounted a (pre-scripted)
stressful experience.
%The operator was tasked with making the robot express empathy, with the
%researchers' aim being to better understand the importance of movement and
%voice in teen-robot interaction. Notably, the teleoperated nature of the robot
%was made explicitly clear to all participants (somewhat at odds with typical
%WoZ studies which tend to deceive the participant with respect to the robot's
%autonomy). 
In a second experiment, they utilised virtual reality (VR) such that one teen
interacted, in an immersive VR environment, with a robot
avatar teleoperated by a teen outside of that VR environment. From this study, the
authors gathered data about the way teens collaborate and their perceptions of
robot roles and behaviours. To this end, they demonstrated the value in expert
(user) teleoperation of a proposed robot, both for better understanding the use
case requirements and user needs, but also as a way to generate exemplars of
desirable autonomous robot behaviour.
\citet{alves-oliveiraYOLORobotCreativity2017} also
demonstrated a similar use of puppeteering and roleplay methods as part of a
co-design study with children.

In summary, work to date has demonstrated how PD methods can be used to study a
proposed application domain in an attempt to ensure researchers thoroughly
understand the context of use, and to elicit some expert knowledge for informing
robot design and automation. This goes some way to supporting a mutual shaping
and responsible robotics approach to social robot development. However, there remains
two key disconnects in delivering truly end-to-end PD and mutual shaping in
development of an autonomous social robot. Firstly, robot automation is informed but not controlled or developed by domain
experts. Secondly, there is a disconnect between this program definition and the
real-world interaction requirements and situational specificities that will
likely be crucial to overall robot success when deployed in real world interaction.

\smallskip

\subsection{Alternative Methods to Capture Domain-Expert Knowledge}

One of the key assumptions of PD in the context of robotics research is that the
knowledge of the desired robot behaviour is held by domain experts and needs
to be translated into programs. Typically, this translation is made by
engineers, obtaining a number of heuristics from the domain experts and
consequently automating the robot. Although widely applied even in PD research,
this method only partially delivers on the promises of PD, as domains experts
are used to inform robot behaviours but still rely on external actors (the
engineers) to transform their intuition, knowledge, and expertise into actual
code. Furthermore, this process can lead to a number of communication issues as members from
different communities have different ways of expressing needs and desires. Nevertheless, there exists a number of alternative
solutions to capture domain-expert knowledge that could support a PD
approach to robot automation.

\smallskip

\subsubsection{End-user programming tools}

A first solution is to create tools to allow domain-experts to create robot
behaviours themselves. Research in end-user development, or end-user
programming, explores tools to allow domain experts or end-users to create
programs without requiring coding knowledge. Typical applications are home
automation, application synchronisation (e.g., IFTTT or Microsoft Flow), or
video games development. Additionally, end-user programming has seen large
interest in robotics, for example to create autonomous robot behaviours for both
industrial robots \citep{paxton2017costar,gao2019pati} and social robots
\citep{leonardi2019trigger,louie2020social}. These \emph{authoring} tools are
often developed by engineers and then provided to users to create their own
applications without
relying on text-based coding, for example by using visual or block programming
\citep{huang2017code3}, tangible interfaces \citep{porfirio2021figaro}, or
augmented reality \citep{cao2019ghostar}. 

However, while being more friendly for users, such methods still suffer from two
main drawbacks. First, the interface is often developed by engineers without
necessarily following principles of participatory design. Second, these methods
often see the programming process as a discrete step leading to a static
autonomous behaviour with limited opportunity to update the robot behaviour or
little focus on testing and evaluating the created behaviour in real
interactions. More precisely, users of these tools might not be the actual
target of the application interaction and would program robots outside of the real
context of use, forcing the aspiring developers to rely on their internal
simulation of how the interaction should happen. For example, a shop owner could
use an authoring tool to create a welcoming behaviour for a social robot, test
it on themselves while developing the behaviour and then deploying it on real
clients with limited safeguards. In such process, developer have to use their
best guess to figure out how people might interact with the robot, and often
have issues to infuse the robot with tacit knowledge, such as timing for actions
or proxemics. This disconnect can lead to suboptimal robot behaviour as the
robot will face edge cases in the real world that the designer might not have
anticipated.

\smallskip

\subsubsection{Learning from Real-World Interaction(s)}

A method to address this gap between an offline design phase and the real world
is to mimic the expert while they perform the interaction. Using machine
learning, systems can learn from the experts how robots should behave. For
example, \citet{liu2016data} asked participants to role play an
interaction between a shopkeeper and a client and recorded data about this
interaction (e.g., participants location or speech). From these recordings, Liu
et al. learned a model of the shopkeeper, transferred it to the robot, and
evaluated it human-robot interactions. Similarly, 
\citet{porfirio2019bodystorming} recorded interaction traces between
human actors and formalized them into finite state machine to create a robot
behaviour. While relying on simulated interactions, these methods provide more
opportunities to developers to explore situations outside of their initial
imagination.

One assumption of these methods is that robots should replicate human behaviour.
Consequently, such methods allow the capture of implicit behaviours such as the timing
and idiosyncrasies of human interactions. However, real-world interactions
with robots might follow social norms different from ones between humans only.
Consequently, learning directly from human-human interaction also presents
limitations.

Wizard of Oz (WoZ) is an interaction paradigm widely used in robotics
\citep{riek2012wizard} whereby a robot is controlled by an expert deciding what
actions the robot should execute and when. The main advantage of this paradigm is to
ensure that the robot behaviour is at all times appropriate to the current
interaction. For this reason, WoZ has been extensively used in Robot-Assisted
Therapy and exploratory studies to explore how humans react to robots. Recent
research has explored how this interaction can be used to collect data from real
human-robot interaction and learn an appropriate robot behaviour. \citet{knox2014learning}
proposed the ``Learning from the Wizard'' paradigm, whereby a robot would first
be controlled in a WoZ experiment used to acquire the demonstrations and then
machine learning would be applied offline to define a policy.  \citet{sequeira2016discovering}
extended and applied this Learning from Demonstration (LfD), with an emphasis on the concept of
``Restricted-perception WoZ'', in which the wizard only has access to the same
input space as the learning algorithm, thus reducing the problem of
correspondence between the state and action spaces used by the wizard and the
ones available to the robot controller. Both of these works could
support a PD approach to robot automation, as they could be used to generate an
autonomous robot action policy based on data from (non-roboticist) domain expert
WoZ interactions in real-world environments. 

%We build on these works to increase the support for PD (and mutual shaping)
%primarily by utilising \textit{online} machine learning (rather than focussing
%on collection of a dataset to be used in offline learning) and by integrating
%this expert training process in to an end-to-end PD process such that the same
%expert(s) can train the robot that \textit{they} have co-designed.

% Nevertheless, a key drawback of such approaches is that once the demonstrations
% are collected and used for learning, limited control can be applied on the
% learned behaviour to ensure that the behaviour is up to the desired standards.
% The only way to evaluate the system is to test it with limited safeguards.
% Furthermore, the training of the system often requires heavy involvement
% from technical experts to analyse the demonstrations, transform them into a
% dataset and apply the learning algorithm.
Nevertheless, the typical WoZ puppeteering setup results in an absence of interaction between the design/development team and the robot, which prevents
designers from having a realistic mental model of the robot behaviour and does not allow for any mutual shaping between the wizard, the robot and the contextual environment. When collecting data through LfD,
it is not possible to know, during the teleoperated data collection process, at
what point `enough' training data has actually been collected, as the system can
only be evaluated once the learning process is complete. Similarly, when using
end-user programming methods there is little opportunity to know how the system
would actually behave when deployed in the real world. This lack of knowledge
about the actual robot behaviour implies that robots have to be deployed to
interact in the real world with limited guaranties or safeguards ensuring their
behaviours are actually efficient in the desired interaction.
\smallskip

\subsubsection{Interactive Machine Learning}

Interactive Machine Learning (IML) refers to a system learning online while it
is being used \citep{fails2003interactive,amershi2014power}. The premise of IML
is to empower end-users while reducing the iteration time between subsequent
improvement of a learning system. Using IML to create robot behaviours through an interaction
between a designer and a autonomous agent allows for full utilisation of the expert's teaching
skills. It has been shown that humans are skilled teachers who can react to a
learner's current performance and provide information specifically relevant to
them \citep{bloom19842}. Similarly, previous research showed that this effect
also exists, to a certain extent, when teaching robots. Using Socially Guided
Machine Learning \citep{thomaz2008teachable} a human teacher adapts their
teaching strategy to robot behaviour and thus helps it to learn better.  If able
to observe (and correct) the robot's autonomous behaviour, seeing the result of
the robot behaviour as it progresses, the expert can create a model of
the robot's knowledge, capabilities and limitations. This understanding of the
robot reduces the risk of over-trusting (both during training and/or autonomous
operation) and introduces the potential for expert evaluation to become part of
the verification and validation process.

\section{A Blueprint for End-to-End Participatory Design}
\label{sec:blueprint}

We identify the following requirements to extend participatory design into an
\emph{end-to-end} methodology, that include the co-design of the robot's
automated behaviour. Such a method needs to allow for:

%To be an end-to-end PD, a design method would require:

\begin{enumerate}
    \item Systematic observation and study of the use case environment in which
        the robot is to ultimately be deployed;

    \item Inclusion and empowerment of relevant stakeholders (users, domain
        experts) from the initial design phases, such that design and
        application of the robot/interaction scenario is co-produced by
        researchers and stakeholders together;

    \item (Safe and responsible) evaluation of prototypes in the real-world
        environment(s) into which the robot is eventually intended to be
        deployed;

    \item Inclusion of relevant stakeholders in creation of autonomous robot
        behaviours, which should utilise interaction data collected in the
        real-world;

    \item Two-way interaction between the expert `teacher' or designer and robot
        `learner' such that the teacher can better understand the state of the
        robot/to what extent learning `progress' is being made and hence adapt
        their teaching appropriately/flag any significant design issues;

    %\item The (ultimately responsible) domain experts (and other stakeholders
        %as appropriate) to be given an opportunity to develop a model of the
        %robot's behaviour and actively refine the robot's behaviour until
        %reaching satisfaction.

    \item Inclusion of relevant stakeholders in (safe) evaluation of autonomous
        robot behaviours, as they perform in the real-world.

    %\item Allow these domain experts to have a model of the robot behaviour.
    %\item Allow domain experts to assess the behaviour in a safe and realistic way.

\end{enumerate}

Requirements 1 and 2 can be addressed by the typical PD methods discussed in
Section \ref{sec:relatedPD}, and requirements 3 and 6 can be addressed by
carefully designed `in-the-wild' studies. In our work, we therefore look to
specifically tackle requirements 4 and 5 by demonstrating how robot automation
can be approached as an in-situ, triadic interaction between domain expert
teacher(s), robot learner and target end user(s). With \meth, we showcase how this approach 
can be integrated into one continuous, end-to-end PD process that satisfies all
of the above requirements.

Table~\ref{tab:blueprint} summarises the key outcomes of, and some potential tools for each stage of \meth.
Figure~\ref{fig:methodcomp} shows how these steps compare to typical PD, as well
as who (domain experts and/or engineers) are involved at each stage. Each stage is detailed in full below. 
Table~\ref{tab:studies} shows how these steps have been derived from/were represented in our two foundational
studies.

% visualises this process as applied for~\cite{winkle2020insitu}'s fitness coach
% robot, as an alternative to the typical PD process presented in
% Table~\ref{tab:typicalPD}.

\begin{table}[]
\caption{Key outcomes of and appropriate tools for each stage of \meth.}
\begin{centering}
\tymin=.3\linewidth
\begin{tabulary}{\linewidth}{LLL}
\toprule
                            & Outcomes & Tools \\ 
\midrule
1. Problem Definition       & Domain understanding 
                            & Ethnographic studies, focus groups, brainstorming     \\
\vspace{.2cm}\\
2. Interaction Design       & Interaction scenario, robot selection/design 
                            & Workshop, role-playing, low-tech prototyping          \\
\vspace{.2cm}\\
3. System Specification     & State-Action space for the robot, teaching tools 
                            & Brainstorming, behaviour prototyping                  \\ 
\vspace{.2cm}\\
4. Technical Implementation & Robot system (sensors and actions), teaching system (authoring tools or learning algorithm) 
                            & Software development, lab studies, testing workshops  \\ 
\vspace{.2cm}\\
5. Real-World Deployment    & Delivering on the application target, autonomous robot 
                            & In-situ teaching by expert                            \\ 
\bottomrule
\end{tabulary}
\end{centering}
\label{tab:blueprint}
\end{table}

\smallskip

\subsection{Step 1: Problem Definition}

As noted in Figure~\ref{fig:methodcomp}, Step 1 of our method aligns to best
practice use of PD as previously demonstrated in social robotics. The
purpose of this stage is to generate a thorough understanding of the use context
in which the robot is to be deployed, and to invite stakeholders to influence
and shape the proposed application. It would likely include observations, focus
groups and/or interviews with a variety of stakeholders.   

The focus group methodology presented in~\citep{winkleMutualShapingDesign2019}
is one appropriate method that could be used for engaging with stakeholders at
this stage as it facilitates expert establishment of non-roboticists, broad discussion of the application context (without presentation of a pre-defined research agenda), participant reflection on the context of use `as is' and researcher-led sharing of technical expertise; followed by detailed consideration and refinement of the research  agenda based on researchers and participants now being equal co-designers.

\smallskip 
%Ideally, such stakeholder engagement would be conducted in conjunction with
%observation and systemic study of the context of use...[example(s)].    

\subsection{Step 2: Interaction Design}

Similarly to Step 1, Step 2 of our method also aligns to best practice use of PD as
previously demonstrated in social robotics. The purpose of this stage is to
define and refine the interaction scenario(s) the proposed robot will engage in,
and hence the functionalities/capabilities it might require. The robot platform
should also be chosen at this stage. For simplicity here we have equated robot
platform \textit{choice} with robot platform \textit{design}. Much current
social robotics research utilises off-the-shelf robot platforms (e.g. Pepper and
NAO from Softbank Robotics) but others focus on the design of new and/or application-specific
platforms. Either can be appropriate for \meth as long as the choice/design
is participatory with stakeholders (for a good example of PD in design of a
novel robot, see \citealt{alves-oliveiraYOLORobotCreativity2017}'s work on designing the YOLO
robot. 

Focusing then on more specific application of the robot and the interaction(s)
it should engage in, methods for PD might include focus groups and similar as
per Step 1, but could also include more novel and/or creative PD activities such
as script writing~\citep{bjorlingParticipatoryResearchPrinciples2019},
roleplaying (including also stakeholder teleoperation of the
robot)~\citep{bjorlingParticipatoryResearchPrinciples2019,alves-oliveiraYOLORobotCreativity2017}
and accessible, `low-tech'
prototyping~\citep{valenciaCodesigningSociallyAssistive2021}. 

Note that there is an important interaction design decision to be made here
regarding what final deployment of the robot `looks like' in terms of long term
oversight by/presence of domain expert(s) (those involved in it's co-design or
otherwise) and the role those experts play with regards to the target user. This can be reflected in the teaching interaction setup, specifically with regards to the amount of interaction between the domain expert(s) and target users (see Figure~\ref{fig:three-way}). %, as per Figure~\ref{fig:autonomy-exper}. 
For
example, it was decided early on in the design of~\cite{winkle2020insitu}'s
fitness coach robot that there was no intention to ever fully remove the expert
presence from the interaction environment. As an alternative, in \cite{senft2019teaching} the intention from the onset was
to create a fully autonomous and independent robot that interacted alone with
the target users. Such decisions regarding the role of domain experts would ultimately emerge
(explicitly or implicitly) in conjunction with deciding the robot's
functionalities and the further system specification undertaken in Step 3.
However, this long-term desired role of the domain expert(s) should be made
clear, explicitly, at this stage, such that it can be reflected in the approach
to program definition.

\smallskip

\subsection{Step 3: System Specification}

As shown in Figure~\ref{fig:methodcomp}, it is at this stage that our method
begins to diverge from the typical PD process, although we continue to utilise
PD methods. This step is concerned with co-design of system specifities required
to (i) deliver the interaction design resulting from Step 2 and (ii) facilitate
expert-led teaching phase on real-world deployment that is fundamental to
our method (see Step 5). In summary, the aim of this step is to co-design the
robot's action space, input space, and the tool(s) required to facilitate the bi-directional teaching interaction between the domain expert and the robot. There is also some similarity here to the design process for a WoZ
or teleoperated system, which would also require design of the robot's action
space and an interface for (non-roboticist) teleoperation of the robot. The key
difference here is the additional requirement to specify the robot's input space
and the choice of teaching tools for the move towards autonomy during Step 5.

% Drawing from ~\cite{winkle2020insitu}'s fitness coach robot as a case study,
% note that:

% \begin{itemize}
%     \item The `teaching' interface (representing our tool for facilitating the expert-robot teaching interaction) was also co-designed (iteratively, with prototype
%         testing being undertaken in-the-wild as per
%         Tables~\ref{tab:studies} and~\ref{tab:storyboard}) with the domain expert who would be using it
%         on deployment.

%     \item The domain expert was entirely responsible for writing specific robot
%         utterances to match those actions he had co-designed, further ensuring
%         his ability to predict how the system would behave on deployment.

%     \item A number of the design activities (mock sessions, prototype
%         evaluations) were conducted in the actual gym environment in which the
%         robot was to be deployed (again depicted in Table~\ref{tab:storyboard}).

%     \item The domain expert was involved in finalising all details regarding the
%         final interaction scenario and proposed deployment, including e.g.
%         physical placement of the robot and himself with respect to
%         participants. 

% \end{itemize}

\smallskip

\subsection{Step 4: Technical Implementation}

The main development effort for our method lies in producing the full
architecture and tools to allow domain experts to specify autonomous robot behaviour. We
note here that the technical implementation required is likely to be greater
than that required for a typical WoZ setup and might not be simpler than heuristics-based robot controller. 

Four main components need to be developed during this phase:

\begin{enumerate}
    \item Set of high-level actions for the robot.
    \item Set of sensory inputs that will be used to drive the future robot behaviour.
    \item A representation of the program which will encode autonomous behaviour.
    \item Expert tools to specify the mapping between the sensory state and the actions.
\end{enumerate}

With our method, the program representation could take the shape of a machine learning
algorithm taking inputs from the expert via the interface and learning a mapping
between the current state of the world when the action was selected and the
action itself (the approach taken in our foundational works). Alternatively, the representation could allow the expert to
encode a program explicitly, for example through state machines, or trigger-action programming, while allowing
the expert to update the program in real time, and control the robot actions to
ensure that they are constantly appropriate.

A typical automation system would replace the expert tools with an actual
definition of the behaviour making use of the program representation to map
sensors to actions and define fully an autonomous behaviour. On the other end of
the spectrum, a WoZ setup might not need a representation of the program, but
instead would rely on the interface to display relevant sensory inputs to the
wizard (if any) and allow them to select what action to do.

\subsection{Step 5: Real World Deployment and Teaching Phase}

Undertaking robot automation (and evaluation) in-the-wild is a key part of \meth. To support a mutual shaping approach to robot design and ensure appropriate robot behaviour, the teaching phase should adhere to the following requirements:

\begin{enumerate}
    \item it must be undertaken in situ, i.e., in the context of the final context of use, and with the real target population.

    \item it must utilise a domain expert teaching the robot as it delivers on the application interaction.
    
    \item the expert-robot interaction should be bi-directional, i.e., the expert should be able to define and/or refine the robot's autonomous behaviour policy, while the robot informs the expert about its status.

\end{enumerate}

Requirement 1 ensures that the approach is ecologically valid, and that the information used by the expert for the automation are suited to the real challenges and idiosyncrasies of the desired context of use. 

Requirement 2 ensures that people with domain knowledge can encode that knowledge in the robot. Furthermore, the presence of the expert should be used to ensure that the robot is expressing an appropriate behaviour at all times. As the teaching happens in the real world, with the real users, there is limited space for trial and error. The expert can be used as a safeguard to ensure appropriate robot behaviour even in the initial phases of the teaching.

Requirement 3 ensures that the expert can create a mental model of the robot behaviour. This point is a key difference to non-interactive teaching methods such as the ones based on offline learning (e.g., \citealt{sequeira2016discovering}). With the robot's feedback on its policy (through suggestions or visual behaviour representation), the expert can assess the robot's (evolving) capabilities and decide what inputs would improve the robot's policy further.

Finally, during this real-world deployment, if the robot is ultimately expected to interact autonomously/unsupervised, the expert can use their mental model of the robot behaviour to decide when enough teaching has been done and when the robot is ready to interact autonomously. By relying on online teaching, this decision does not have to be final as the expert could seamlessly step back into the teacher position when the robot interacts with sensitive populations or if the robot requires additional refinement of its policy.

%\begin{figure}
%  \centering
%  \includegraphics[width=\linewidth]{FlowChartDraft}
%   \caption{An end to end methodology for participatory design of autonomous social robots. Note that the use of online teaching for program definition essentially allows steps 5 and 6 to be merged and conducted in parallel. Alternative methods for program definition and user testing might be used, but online teaching specifically suits situations where (i) the (human expert) reasoning for particular action selection/evaluation is highly intuitive, (ii) the context of use requires continuously appropriate/effective robot behaviour. \katie{boxes are long steps, so keep 5 and 6 as they are on the left; the right side is some other chart shapes to represent the instantaneous nature of these feedback, and highlight that we can do 5 and 6 at the same time. Could colour code what is engineer and what is domain expert and what is both - note that right hand side is *only* domain expert + robot. Change the diagram such that it's left side VS right side rather than right side == left side.}}
% \caption{...}
%  \label{fig:method}
%\end{figure}

% \section{Case Study: Robot Fitness Coach}
% \begin{itemize}
%     \item story board of process as applied to design of the C25K gym coach [K] \katie{I already presented a table with a description of the pre-study co-design sessions in the RSS paper - is it ok to replicate that here? (I think it would be useful in showing exactly *how* you might implement that co-design of the base system)}
% \end{itemize}

\section{Foundational Studies}

The \meth method is primarily derived from two foundational studies made by
the authors, which were themselves informed by the authors' previous experiences
working with domain experts in the design of social robots. The
first one, presented in \citet{senft2019teaching}, explores a study with 75
children how the teaching interaction could be used to create an autonomous
robot behaviour. As shown in Table \ref{tab:studies}, this study did not employ
PD, the authors, researchers in HRI, did the early steps by themselves based on
their previous related experiences. The second one, presented in~\citet{winkle2020insitu}, built on
the first study by utilising the same teaching approach to robot automation, but
incorporating that into and end-to-end participatory design process to support
mutual shaping. The end goal of each study was also slightly different, as
\citet{senft2019teaching} aimed to produce a robot that would ultimately
interact with users with little-no further expert involvement.
\citet{winkle2020insitu} also aimed to produce an autonomous robot that would
primarily interact 1:1 with users, but with no desire to remove the expert, who
would have their own interactions with the users, and/or provide provide
additional teaching to the robot should they deem it necessary. 

\begin{table}[]
\begin{center}
\tymin=1.5cm
\begin{tabulary}{\linewidth}{LLL}

\toprule
       & School Based Educational Robot & Gym Based Robot Fitness Coach \\ 
\midrule   
Step 1  & Decision by researchers based on experience to focus on learning food chain around an educational game for children of age 8-10.
        & Researchers identified the NHS C25K exercise programme based on research goals (longitudinal, real-world HRI) but worked with a fitness instructor to observe typical environment and refine problem definition.\\ 
\vspace{.1cm}\\
Step 2  & Decision by researchers to focus on robot-user interaction, with expert only providing robot commands and oversight of the robot behavior to ensure that each action is validated by them. Goal is to evaluate the creation of an autonomous robot.
        & Decision in conjunction with the fitness instructor that the robot would lead exercise sessions (in which he would minimise interaction with exercisers) but that he would provide additional support (e.g. health advice, stretching) outside of these. Goal is to create and demonstrate an effective, real-world SAR based intervention via PD (as responsible robotics). \\  
\vspace{.1cm}\\
Step 3  & Using SPARC \cite{senft2015sparc} as interaction framework, robot state and action spaces defined by researchers. Teaching through a GUI on a tablet.
        & Also using SPARC~\cite{senft2015sparc} the robot state and action spaces as well as the teaching GUI were all co-designed with the fitness instructor.\\ 
\vspace{.1cm}\\
Step 4  & Implementation of all the actions and learning algorithm. Prototype evaluation in lab. Initial pilot study in schools for evaluating the game with the target population and used as teacher training.
        & Implementation of all the actions and learning algorithm. Fitness instructor provided all dialogue for robot actions. Prototype evaluation was undertaken in the lab and in the final study gym environment, final robot placement and system installation details were also decided in conjunction with the fitness instructor.\\ 
\vspace{.1cm}\\
Step 5  & Deployment in two local schools with more than 100 children over multiple months. Between-subject evaluation with three conditions: a passive robot, a supervised robot (during the teaching interaction), and an autonomous unsupervised robot.
        & Deployed in to the university gym for teaching and autonomous evaluation through delivery of the C25K programme (27 sessions over 9-12 weeks) to 10 participants. Ran a total of 232 exercise sessions of which 151 were used for teaching the IML system, 32 were used for evaluating the IML system when allowed to run autonomously and 49 were used to test a heuristic-based `control' condition (all testing was within-subject).\\ 
\bottomrule
\end{tabulary}
\caption{Overview of activities undertaken in the two case studies as exemplars for applying our generalised methodology. See Table~\ref{tab:storyboard} for a pictorial `storyboard' of this process and the co-design activities undertaken for development of the Robot Fitness Coach.}
\label{tab:studies}
\end{center}
\end{table}

\smallskip

\subsection{Study 1: Evaluating the Teaching Interaction}

The goal of this first study was to evaluate if the teaching interaction could
be used to create autonomous social behaviours \cite{senft2019teaching}. This
study was designed by the authors, who had experience designing robots for the application domain, but did not involve external stakeholders such as teachers.

During the problem definition phase, researchers decide to contextualize the
work in robot tutoring for children, and explore questions such as how robots
can provide appropriate comments to children (both in term of context and time)
to stimulate learning. This work was based on experience and knowledge from the
researchers about educational robotics. 

During the interaction design phase, researchers decided to focus the
application interaction around an educational game where children could move
animals on a screen and understand food nets. This part included an initial
prototype of the game. As the goal was to explore how autonomous behaviours
could be created, the teacher was not involved in the game activity, only the
robot was interacting with the child. The robot used was a NAO robot from
Softbank Robotics. 

In the system specification, the state and action spaces of the interaction were
selected. Examples of state include game-related component (e.g. distance
between animal) and social dynamics elements (e.g. timing since last action of
each agents). The robot's actions were divided into five categories:
encouragements, congratulations, hints and reminding rules.
%and ?. 
The teacher-robot interaction used SPARC \cite{senft2015sparc}.

In the technical implementation phase, the learning algorithm was developed,
tested and interfaced with the other elements of the system. The teaching
interface was also created in such a way as to allow the teacher to select actions for the robot
to execute and receive suggestions from the robot. At this stage, initial
prototypes were tested in lab studies and schools.

In the real-world deployment, authors evaluated the system in two different
schools with 75 children. The study adopted a between-participant design and
explored three conditions: a passive robot, a supervised robot (referring to the
teaching interaction), and an autonomous robot (where the teacher was removed
from the interaction and the learning algorithm disabled).

Results from the study showed that the teaching interaction allowed the teacher 
to provide demonstrations to the robot to support learning in the real world. 
The teacher used the teaching interaction to create a mental model of the robot
behaviour. When deployed to interact autonomously, the robot enacted a policy
presenting similarities with the one used by the teacher in the teaching phase:
the frequency of actions was similar and the robot captured relation and timing between
specific events and actions (e.g. a congratulation action should normally be executed
around two seconds after an eating event from the child's actions). Overall, this
study demonstrated that human can teach robot social policy from in-situ guidance.
 
\subsection{Study 2: Teaching Interaction as Participatory Design}

The goal of this study was to use the teaching interaction approach to
facilitate creation of a fully expert-informed/expert-in-the-loop autonomous socially
assistive robot-based intervention for the real-world. The fundamental activity
to be delivered by robot, the NHS C25K programme, was selected by the
researchers based on this research goal; but all study implementation details
were decided and designed in conjunction with a domain expert (fitness
instructor) throughout. Given the end-to-end and constant expert involvement for
this study there was seamless progression and some overlap between the problem
definition, interaction design and system specification phases as we present them
for \meth. A number of co-design activities were undertaken
(over a total of 6 sessions totalling approx. 12.5 hours) which ultimately
covered all of these key phases, sometimes in parallel, allowing for iteration
of the overall study design. 

Problem definition was achieved by researchers working with the fitness
instructor to (i) understand how a programme like C25K would be delivered by a
(human) fitness instructor and (ii) explore the potential role a social robot
might take in supporting such an intervention. This involved the researchers
visiting the university gym and undertaking mock exercise sessions with the
instructor, and the instructor visiting the robotics lab to see demonstrations
of the proposed robot platform and a presentation by the researchers on their
previous works and project goals. The robot used was a Pepper robot from
Softbank Robotics. 

For the interaction design, the researchers and fitness instructor agreed that
exercise sessions would be led by the robot and primarily represent robot-user
interactions, with the fitness instructor supervising from a distance and only
interacting to ensure safety (e.g. in the case of over exertion). As this study
also aimed to test (within-subject) the appropriateness of resultant autonomous
behaviours, it was decided to purposefully leave the details of the fitness
instructor's role somewhat ambiguous to exercising participants. The instructor
was not hidden away during the interaction and it was clear he was supervising
the overall study, but exercisers were not aware of the extent to which he was
or wasn't engaging in teaching interactions with the robot during sessions. As noted in
Section~\ref{sec:blueprint}, deciding on what long term deployment should `look
like' in terms of robot-user-expert interactions is a key design requirement at
this stage. For the robot fitness coach, we imagined a `far future' scenario,
where one of our robot fitness coaches would be installed next to every
treadmill on a gym floor, supervised by one human fitness instructor. That
instructor would ensure exercisers' physical safety and still play a role in
their motivation and engagement as human-human interaction is known to do. This
type of interaction with one expert, multiple robots and multiple target users
is a common goal in many assistive robot  applications where some tasks could be
automated, but there is a desire to  keep an expert presence to e.g. maintain
important human-human interactions and ensure user safety.

% We imagine the fitness robot coach(es) initially being taught by the fitness
% instructor via 1:1 supervised sessions such that they could teach the robot
% system how to act most effectively for each individual exerciser (as per our
% experimental setup in~\cite{winkle2020insitu}). Then, we imagined said fitness
% instructor providing continual overall supervision when the robot coach(es) are
% working simultaneously with multiple users, and providing additional training
% data to the robot coach(es) again where necessary, e.g. in case of new
% exercisers joining the gym or in case of an existing exerciser's needs changing
% over time.  

The system specification represented somewhat of a `negotiation' between the
researchers and the fitness instructor, as he identified the kind of high level
action and inputs he felt the robot ought to have, and the researchers
identified how feasible that might be for technical implementation. The state
space consisted of static and dynamic features that were designed to capture
exerciser engagement, task performance and motivation/personality; all
identified by the fitness instructor as being relevant to his decisions in
undertaking fitness instruction himself and hence teaching the robot how best to
interact with a particular participant. The action space was divided into two
categories: task actions and social supporting actions. The task actions were
fundamentally set by the C25K programme (i.e. when to run or walk and for how
long at a time). The social supporting actions were then broken down into eight
sub-categories covering time reminders, social interaction, praise, checking on
the user, robot animation and proxemics (leaning towards/away from the user).
%Task and social supporting actions could then be `styled' postive, challenging,
%negative or neutral...  
Importantly, system specification for this study also included co-designing the
GUI that would facilitate the bi-directional teaching interaction (also
utilising SPARC \cite{senft2015sparc}) between the robot and the fitness
instructor with the fitness instructor himself. 

The technical implementation phase essentially mirrored that of Study 1: the
learning algorithm was developed, tested and interfaced with the other elements
of the system. The teaching interface was also finalised based on the co-design
activities described previously, and similarly allowed the fitness instructor to
select actions for the robot and to respond to its suggestions. Initial
prototypes of both the robot and the GUI were tested in lab studies and the
final gym environment.

In the real-world deployment, researchers evaluated the system in a university gym
with 10 participants recruited to undertake the 27-session C25K programme over a
maximum of 12 weeks. The study adopted a within-subject design and explored
three conditions: a supervised robot (referring to the teaching interaction), an
autonomous robot (where the fitness instructor was still in position but allowed
all learner-suggested actions to auto-execute) and a heuristic-based autonomous
robot; a `control' condition for comparing the `teaching interaction as PD'
approach to, representing a `next-best' alternative for generating
expert-informed autonomous behaviour. 

Results from the study again demonstrated the feasibility of SPARC and IML for
generating autonomous socially assistive robot behaviour, suggested that the
expert-robot teaching interaction approach can have a positive impact on robot
acceptability (by the domain expert and targets users) and that the teaching
approach yields better autonomous behaviour that expert informed heuristics as a
`next best' alternative for expert-informed autonomous behaviour creation.

\subsection{Evidence of Mutual Shaping}

Typical PD facilitates mutual shaping as it allows non-roboticist, domain experts to shape research goals, design guidelines, and evaluate robot prototypes etc. Here, we reflect on observations of mutual shaping effects in our foundational works, specifically resulting from our teaching approach to robot automation.

During our first study, we observed evidences of mutual shaping and the teacher creating a
mental model of the robot. For example, our teacher realised with experience
that children tended have issues with some aspect of the game (i.e., what food a
dragonfly eats). Consequently, she changed her strategy to provide additional
examples and support for this aspect of the game. Similarly, the teacher also
found that the robot was not initiating some action often and consequently used
some actions more frequently toward the end of the teaching phase to ensure that
the robot would exhibit enough of these actions. This exactly evidences that notion that human teachers can tailor their teaching to a (robot) learner's progress \citep{bloom19842,thomaz2008teachable}

In the second study, we were able to demonstrate mutual shaping in the way the fitness instructor used the robot differently for different participants and/or at different stages of the C25K programme. The longitudinal nature of this study, combined with our approach in supplementing the diadic robot-user interactions with expert-user interactions, meant the fitness instructor got to know each user's exercise style/needs and could tailor the robot's behaviour accordingly. This resulted in the autonomous robot similarly producing behaviour that varied across participants. Similarly, as the programme progressed, the fitness instructor could tailor the robot's behaviour to reflect the changing exercise demands (e.g. using less actions when the periods of running were longer). The flexibility of our approach was also demonstrated when, in response to this increase in intensity, the fitness instructor requested we add a robot-led cool-down period to the end of each exercise session. This was relatively simple to implement from a technical perspective (an additional `walk' instruction at the end of each session plan) but represented a new part of the session for which there existed no previous training data. As we made this change within the teaching phase (before the switch to autonomous operation) the instructor was able to address this, such that the robot was able to successfully and appropriately support this new cool-down phase when running autonomously. 

We also saw an interesting, emergent synergy in the way that the fitness instructor utilised and worked alongside the robot coach. Towards the end of the study, as exercise sessions became more demanding, the fitness instructor took more time at the end of each session to undertake stretching exercises with each participant. This lead to small amounts of overlap between each participant, at which point the fitness instructor would start the next participant warming up with the robot, whilst he finished stretching with the previous participant. We find this to be compelling evidence of the way domain experts will change their practice and/or the way they utilise technological tools deployed into their workplace, particularly when they can be confident in their expectations of how that technology will perform, as is particularly fostered by our approach. 
\smallskip

\subsection{IML for the Teaching Interaction: Opportunities and Limitations}

As noted previously, both of our foundational studies utilised IML via the SPARC paradigm to facilitate the teaching interaction. From a technical perspective, our foundational studies demonstrate the feasibility and relative effectiveness (in terms of teaching time) of this approach. Fundamentally, \meth is agnostic with regards to the specifical computational approach to facilitating the teaching interaction, but we find IML to be a particularly compelling solution, in-line with the overall aims of the method, as it makes for an intuitive bi-directional teaching interaction for the domain expert. Specifically, through one single interface, they can see what the robot intends to do (and potentially why) before that action is executed, improving their understanding of the robot's learning progress, and instantiate teaching exemplars in real-time, informed by that understanding as well as the instantaneous requirements of the application task. 

However, here we draw attention to one key limitation here regarding expert-robot interactions and assessment when using IML. An important element of mutual shaping not considered here is if/how/to what
extent the suggestions made by the learning robots may have influenced the
domain experts. For example, had the learning robots not been making
suggestions, such that the robot was entirely controlled/teleoperated by the
experts, would the action distribution and timing of actions remained the same?
Further, if the experts did not have the ability to actively reject suggestions
(indicating that the learner was not producing appropriate robot behaviour)
would they still have post-hoc identified those actions as being inappropriate? 

This is particularly interesting given the high number of suggested actions
still being rejected at the end of the training phase, in both of our foundational studies, immediately followed by seemingly appropriate robot behaviour that was
positively evaluated by the experts themselves during autonomous operation.
Success of our approach inherently assumes that the domain expert/system
`teacher' would provide a `correct' and fairly consistent response; i.e. that
they (i) can correctly assess the quality of each action suggested by the robot and make an informed about whether this action should be executed and (ii) are always able to ensure that required robot actions are executed in a timely fashion. With SPARC, these robot suggestions are the main mean to help the expert create a mental model of the robot behaviour. Consequently, whilst our results
demonstrate the IML does fundamentally `work' for automating robot behaviour, and that our domain experts did construct a mental model of the robots' behaviour, there remains an open question regarding how the robot could improve the transparency of its behaviour to actively support mental model creation for the teacher.
%these questions are directly linked to its efficacy and thus warrant further
%investigation.

%more exploration about best way to execute the IML / interface and input/output

\section{Discussion}

\subsection{A Flexible and Effective Method for Automating Social Robots}

\begin{figure}
 \centering
 \includegraphics[width=0.8\linewidth]{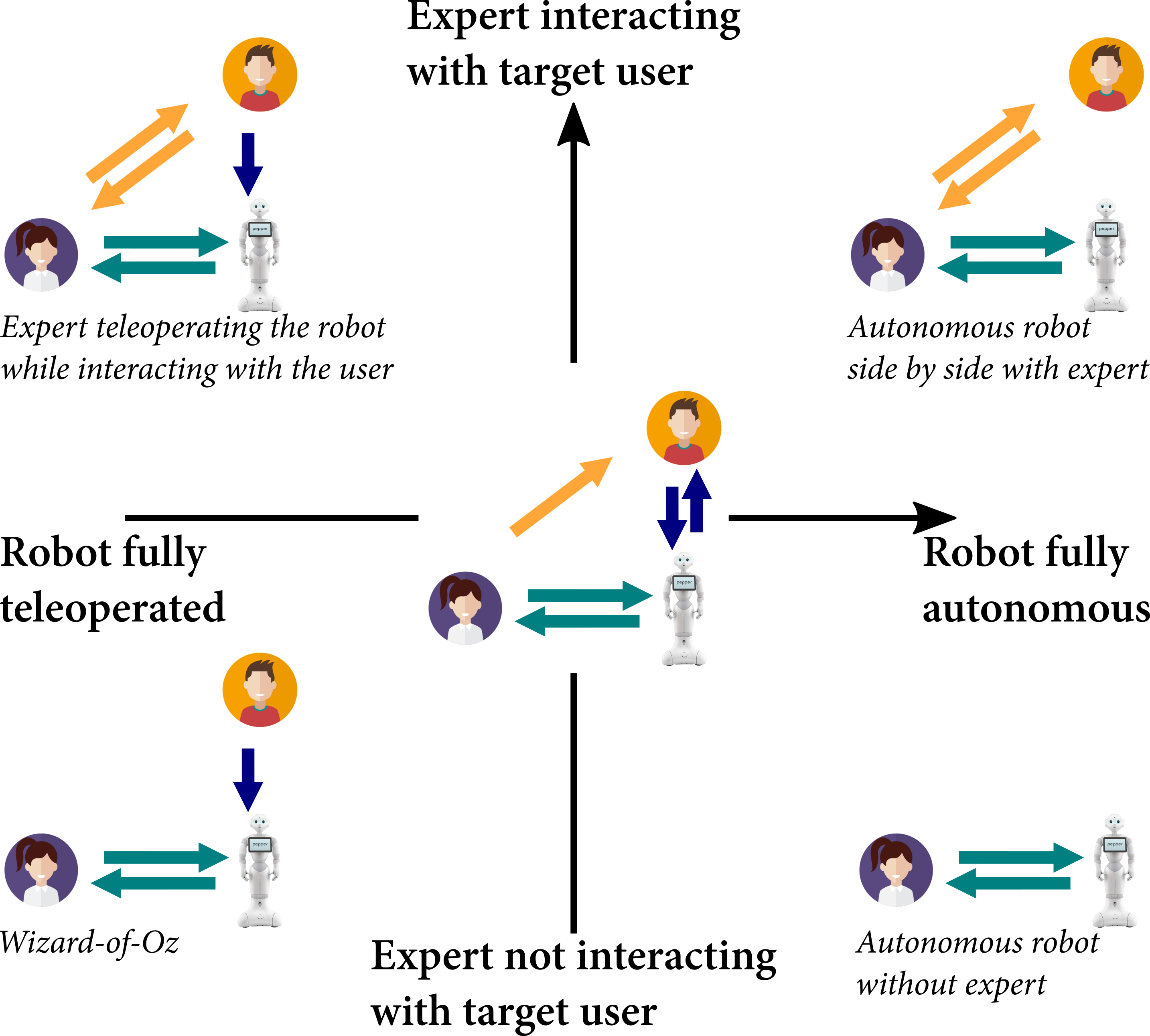}
\caption{Two dimensional representation for visualising the different types of
long-term expert-robot-user interactions that a social robot might be designed
for, all of which \meth can facilitate. Note this is not a discrete space,
and \meth specifically makes it possible to move along these axes upon
real-world deployment.}

 \label{fig:autonomy-exper}
\end{figure}

We suggest that \meth can be used to design robots for a variety of interaction settings, in terms of the required autonomy and the nature of expert-robot-user interactions long-term. We propose two axes to describe the different types of interaction that might be desired, based on the application (Figure~\ref{fig:autonomy-exper}). A first axis describes the extent to which the domain expert(s) and user(s) are expected to interact long term, as a supplement to the robot-user interaction(s). The second axis reflects the autonomy of the robot, from full supervision (teleoperation) to full autonomy. These two axes are independent as, for example, cases exist where the expert might be continuously interacting with the target users, while continuing or not to supervise and/or improve the robot's autonomous behaviour long term. In addition, these axes do not represent a discrete space, as the teaching interaction element of \meth specifically makes it possible to move along either axis at any point during real world deployment.

The robots developed in our foundational studies demonstrate this flexibility, and exist in slightly different spaces on these axes. \cite{senft2019teaching}'s educational robot is an example of an autonomous robot operating without the expert, and the teaching interaction represented a typical Wizard-of-Oz setup, i.e. there was no interaction. \cite{winkle2020insitu}'s robot fitness coach is closer to an autonomous robot operating side-by-side with the domain expert, and the teaching interaction utilised some interaction between the expert and the users (although this was undertaken \emph{outside} of direct teleoperation).

%This three-way interaction between the
%expert, robot, and user can be designed to reflect the long-term aspirations
%for the robot's deployment in its specific context of use.

The two foundational studies also demonstrate different, complimentary elements of the effectiveness of \meth for designing social robots. \cite{senft2019teaching} fundamentally demonstrated the practical feasibility of the teaching interaction for creating appropriate autonomous behaviour. After a teaching phase with 25 children, the robot was deployed autonomously and without expert supervision. It displayed a similar policy to when it was supervised, for example, capturing connections between some events and actions with appropriate timing.

%\subsubsection{End-to-End PD as a Path to Autonomy for Unsupervised Interactions}

% One of the way to use our method is to create fully autonomous robots for use
% in diadic interactions (i.e. with no further expert-robot or expert-user
% interactions). As we have shown in \cite{senft2019teaching}, online teaching can
% be used to reach full autonomy. In this study, after a teaching phase with 25
% children, the robot was deployed without supervision to interact with 25 other
% children.  During this autonomous deployment, the robot policy was fixed, the
% robot was not learning and the teacher could not provide any further input to
% the system or preventing the robot to execute actions. Results from the study
% showed that when the robot was autonomous, it was displaying a similar policy to
% when it was supervised, for example capturing connections between some events
% and actions with appropriate timing.

%\subsubsection{Supplementing Meaningful Human-Human Interaction and Positively Impacting Robot Acceptability in Human-Centred Domains}

Whilst \cite{winkle2020insitu} again demonstrated similarity between supervised and autonomous behaviour, we also specifically demonstrated that the teaching interaction resulted in a better autonomous robot than an expert-informed heuristic based alternative. In addition, we specifically explored to what extent the overall \meth could support mutual shaping and influence robot acceptability. To this end, the significant co-design work undertaken
by the domain expert seems likely to have contributed to the high level of
ownership he seemed to feel toward the system, and the way in which he
conceptualised the robot, throughout, as an independent agentic colleague he was
training. When asked whether he perceived Pepper as more of a tool or a
colleague, the fitness instructor commented \textit{``It was definitely more of a
colleague than a tool...I like to think her maybe early bugs or quirks
definitely gave her a bit more of a personality that maybe I held on to''}. In
addition, when evaluating the robot's performance, the instructor also reflected
on the difference between how the robot might behave in comparison to himself:
\textit{``Pepper's suggestions might not be what *I* would say in that exact
same situation, however it doesn't mean that what was said or suggested was
wrong''}. This gives credibility to the suggestion that \meth can be used to
create robots that do not simply attempt to imitate or replicate the domain
expert directly, but instead play a distinct but complimentary role alongside that domain expert in delivering an assistive intervention. 

The fitness instructor's feedback also suggested use of the robot did not
prevent him from still developing a working relationship with the exercisers,
nor from having a positive impact on their motivation, as he \textit{``did care
about their progress and their health''}. This appears to be true on the
exerciser's side too, as their evaluations suggested they perceived the fitness
instructor and the robot as playing distinct but complimentary roles in their
undertaking of and engagement with the prescribed exercise programme:
\textit{``Pepper was a good instructor and positively motivated my runs. The
    role of Don [the fitness instructor] assisted this in that having him there meant I could
follow the robot’s instructions safe in the knowledge that there was some
support there should anything go wrong!''}

To this end, the fitness coach robot example demonstrates how \meth 
seemingly contributes to robot acceptability, by both domain experts and target
users, and can successfully facilitate meaningful triadic (domain expert - robot
- user) interactions in human-centered domains where there might be a desire to
reduce domain expert workload without ever removing them from the interaction
completely.

\subsection{Supporting `Responsible by Design' Robotics}

The Foundation for Responsible Robotics (FRR) defines responsible robotics as
`the responsible design, development, use, implementation, and regulation of
robotics in society'\footnote{https://responsiblerobotics.org/}. Concerning
research and development, the FRR demonstrates significant overlap with the goals of mutual shaping, and hence our goals in proposing \meth: 

\textit{``Responsible robotics starts before the robot has been constructed.
Ethical decision-making begins in the R\&D phase. This includes the kind of
research practices that are employed; ensuring that a diverse set of viewpoints
are represented in the development of the technology; using methods of
development and production that are sustainable; and taking into consideration
the impact that the technology will have on all stakeholders to mitigate harm
preemptively rather than after the fact.''}

%Clearly then, the focus on stakeholder engagement and participatory design (key
%to the mutual shaping approach) means \meth also represents responsible robotics
%practice. 

A significant number of attempts to more formally define the ethical
design and development have taken the form of published principles of AI and
robotics
\footnote{http://alanwinfield.blogspot.com/2019/04/an-updated-round-up-of-ethical.html},
many of which similarly identify the importance of engaging (non-roboticist)
users and domain experts in robot design and evaluation processes. Arguably one
of the more practical resources is the British standard BS8611-2016 \textit{Guide to the ethical
design and application of robots and robotic systems}~\citep{bsiBS861120162016},
which explicitly identifies ethical risks posed by robots, mitigating strategies
and suggested methods for verification and validation. Notably, the standard
suggests that a number of the identified ethical hazards might be verified and
validated through \textit{expert guidance} and \textit{user validation}. Through
\meth, such guidance and validation is inherently `built-in' to the design
and development process. Based on this, we posit that, in supporting a mutual
shaping approach to robot development, and specifically by `opening up' robot
automation to non-roboticists (such that they can contribute to robot design and
automation, but also better understand robot capabilities and limitations) \meth also represents a concrete implementation of a \emph{responsible robotics} approach, and offers a practical way to
create social robots with expert guidance and user validation being inherent to
the development process.

\subsection{Future Development}

\subsubsection{Inclusion of Application Targets in Design, Automation and Evaluation}

A key limitation in both of our foundational works was the lack of including
target users during the design processes. This is partly because both of
these works concerned the development of robots that would be assisting the 
domain expert practitioners (i.e. a teaching assistant and a fitness
instructor), and so it made sense to focus on working with such experts as
co-designers of the system. However, as discussed in the introduction, inclusion
of all stakeholders is a key aim of mutual shaping approaches to robot
design/development. 

A desire to include target users in the robot’s design and evaluation would raise
the interesting question of how target users, who are expecting to beneficiaries of the interaction, could design the robot. In a number of situations where the 
robot is expected to provide support or additional knowledge, including target
users in the co-design of the action state for example could be either complex
or negate the \emph{illusion} of the robot as an agent. %, potentially negatively impacting on its credibility~\citep{winkleAssessingAddressingEthical2021}.
%in theory are using the robot precisely
%because they themselves are not good at knowing/implementing what they need to
%stay motivated, can make the most useful input to these processes. For example,
%including users in initial detailed co-design of e.g. encouraging actions may
%not be appropriate, but 
However, target users could certainly be included in preliminary testing
of those actions designed with a domain expert. 

\smallskip

% This also raises a number of interesting research questions regarding e.g. if, how and whether
% having one end user act as the expert-in-the-loop training the robot for another
% end user might actually impact on that end user’s own self-understanding and
% motivation to engage with the task. The impact that training a robot via the IML
% process might have on the human-in-the-loop providing that training data is
% another aspect of mutual shaping that could be considered in more detail in
% future works.

\subsubsection{Alternative Teaching/Learning Interactions}

The method presented in this paper focused on a teaching phase where an expert teaches the robot how to interact with a target user, with the target user unaware of the extent to which the expert is involved in the robot behaviour. However, our method is also suited to other interaction designs not explored in our foundational studies.

%target as expert
While situations such as therapy or education require the expert and target user to be different
persons, a large number of other domain relaxes this constraint. For example, an elderly at home 
could have a robot carer, and teach the robot how to support them in their daily activity. In this case, 
the target user is the person knowing best their needs and as such would be the perfect expert. 
\meth would be highly applicable to this situation as the target users could be involved early in 
the design process, help specify the state and action spaces and tools they would need and finally 
teach in situ their robot how to interact while benefiting from the interaction themselves.

Alternatively, building on the previously noted limitation regarding target user inclusion, applications where the robot is to play more of a \emph{peer} role, rather than a expert authority might be best achieved by having one target user teach the robot how to interact with another target user. This might be particularly appropriate for e.g. allowing teenagers to automate companion robots that supporting teenagers' mental health~\citet{bjorlingParticipatoryResearchPrinciples2019}. This raises a number of interesting research questions regarding how the teaching interaction might impact on the teacher's (self-)understanding of the application domain, representing another aspect of mutual shaping that could be considered in more detail in future works. 

%open teaching
An alternative, exciting teaching interaction is having the teaching phase being open and transparent to 
the target user. Teaching robots could be similar to how adult teach children to interact, by providing explicit feedback guideline openly in the social environment. This situation raises a number of open questions such as to what extent having the expert providing feedback to the robot could impact the ascribed agency of the robot, or how could the target user be included in telling the robot how best to help them. We have good evidence from our work~\citet{winkle2020insitu} that such open interaction would not `break the illusion' of the robot being an independent (credible) social agent. Further, previous work suggests that robot users value the human developers `behind' the robot, as it is their `genuine intentions' that underlie the robot's social and assistive behaviours~\citet{winkleEffectivePersuasionStrategies2019}. In sensitive application environments such as the previously mentioned teenage mental health support, such openness may indeed be crucial to robot effectiveness and acceptability~\citet{bjorlingExploringTeensRobot2020}. 

%you have user providing real time feedback to the controlling expert teacher/have the expert teacher asking questions of the target user without it too much 'breaking the illusion' of the robot being an independent (credible) social agent (good evidence this wouldn't be the case from my work + elin's on teens and lack of deception in woz but can't say for sure)
%-> if you were helping the (fitness) instructor make the best (robot) instructor for you, what would you tell them? would it ruin the interaction

However, these alternative teacher/learner configurations need to account for the existing practical constraints of  using reinforcement learning (RL) in human robot interaction. Indeed, in the context of human-robot interaction, RL faces two main issues, (1) the large number of datapoints required to improve the policy (which have to come from real world interaction) and (2) the risks posed by the RL `exploration' in the real human-robot interaction, where the RL algorithm might suggest actions that are inappropriate in a given context.

In our two studies, the domain expert also acted as a `gate keeper' for the robot's suggestions, and as a general safety net, able to intervene if the autonomous robot behaviour was inappropriate. Likewise, when applying \meth in other scenarios, adequate safeguarding needs to be in place, until further research on reinforcement learning can provide adequate safety guarantees. Alternatively, the expert could serve early on to help create an initial safe and effective policy by providing a high amount of guidance. Then, in a second phase, the expert could revert only to the `gate keeper' role, working as a safeguard to ensure that the robot's policy has a minimum efficacy, while letting the robot self-improve. Finally, when the robot reaches a sufficient expertise in the interaction, it could be left to fine tune its policy with less supervision. 

\section{Conclusion}

In this article we present \emph{\method} (\meth), a method for end-to-end participatory design (PD) of autonomous social robots that supports a mutual shaping approach to social robotics. This general method is derived from two, independent foundational studies and represents a culmination of the authors' experiences of working with domain experts in the development of autonomous socially assistive robots. We describe the activities undertaken in those studies to demonstrate how the method has been derived and give tangible examples of how it might be applied. Together, we suggest these foundational studies also demonstrate both the feasibility and the value of the approach, as both resulted in acceptable, autonomous and effective socially assistive robots successfully utilised in complex real world environments.

The first key contribution of \meth is to make robot \emph{automation} participatory, such that non-roboticist, domain experts can contribute directly to generating autonomous robot behaviours. This particularity compliments more typical use of PD for e.g. generating initial robot design guidelines or evaluation robot prototypes. We achieve this expert-led automation by utilising a \emph{teaching interaction}, whereby the domain expert(s) can directly define and refine the robot's autonomous behaviour through a teaching interface. Both of our foundational studies utilised interactive machine learning and the SPARC paradigm~\citep{senft2015sparc} which we suggest is particularly well suited to the overall method goals, therefore we particularly reflect on this approach and its benefits, challenges and limitations. However, whilst we refer to this as a teaching interaction, as the domain expert is `teaching' the robot how to behave, our method is agnostic as to the specific technical approach taken (e.g. machine learning, authoring) to facilitate it.

The second key contribution of our \meth is to facilitate a mutual shaping approach throughout robot development. This is achieved, firstly, by the increased domain expert participation in robot automation as described above. In addition however, our integration of the teaching interaction into real world robot deployment means that this automation of robot behaviour can actually be informed by and reflect the complex and nuanced realities of the real world context, capturing the expert's tacit and intuitive responses to real-world social dynamics. Given that teaching can be re-convened at anytime, the method also facilitates the updating of robot behaviours in response to these dynamics evolving, or new dynamics emerging, i.e. observation of mutual shaping effects. More generally, the in-situ robot deployment and expert teaching role maximises the opportunity to identify and understand such mutual shaping effects to better evaluate the robot's overall impact and efficacy for the proposed application. 

In facilitating end-to-end PD and mutual shaping, we also suggest our method inherently supports responsible robotics, by design. Specifically, it allows for a diverse set of viewpoints to be represented in the development of the technology, and for preemptive consideration of the impact that technology will have on stakeholders. Finally, on a practical level, we also suggest our method can better facilitate multi-disciplinary working as it systematically combines participatory design and technical development such that non-roboticist researchers and stakeholders are no longer excluded from any stage of the development process.   

In summary, we suggest \meth is an all-around effective approach for creating socially intelligent robots, as \emph{practical} as it is \emph{responsible} in facilitating the creation of expert-informed, intuitive social behaviours. We identify a number of areas for potential future development, which we hope will be of interest to other roboticists in refining the method further, and working further towards democratisation of robot design and development. 

\section*{Conflict of Interest Statement}
%All financial, commercial or other relationships that might be perceived by the academic community as representing a potential conflict of interest must be disclosed. If no such relationship exists, authors will be asked to confirm the following statement: 

The authors declare that the research was conducted in the absence of any commercial or financial relationships that could be construed as a potential conflict of interest.

\section*{Author Contributions}

KW and ES led foundational studies 1 and 2 from which this work is derived, both of which were (independently) conducted in close collaboration with SL. KW and ES led on derivation of the generalisable method based on their shared experiences, with all authors contributing to reflections on the foundational studies, resultant implications for the generalisable methodology and producing the final manuscript.

\section*{Funding}
This work was partially supported by the EPSRC via the Centre for Doctoral Training in Future Autonomous and Robotic Systems (FARSCOPE) at the Bristol Robotics Laboratory, University of the West of England and University of Bristol (grant number EP/L015293/1), partially supported by KTH Digital Futures Research Centre, partially funded by the EU FP7 DREAM project (grant no. 611391). 

\section*{Acknowledgments}
We wish to acknowledge our PhD supervisors, Paul Bremner, Praminda Caleb-Solly, Ute Leonards, Ailie Turton, Tony Belpaeme, and Paul Baxter with whom we collaborated on the foundational studies and previous experiences that informed this work. In addition, we wish to acknowledge the two domain experts, Madeleine Bartlett and Donald Knight, who took part in our foundational works whose reflections contributed to our refinement of the method.

\bibliographystyle{plainnat} 
\bibliography{biblio.bib}

\begin{landscape}

    \begin{figure}
        \begin{center}
            \includegraphics[width=0.8\linewidth]{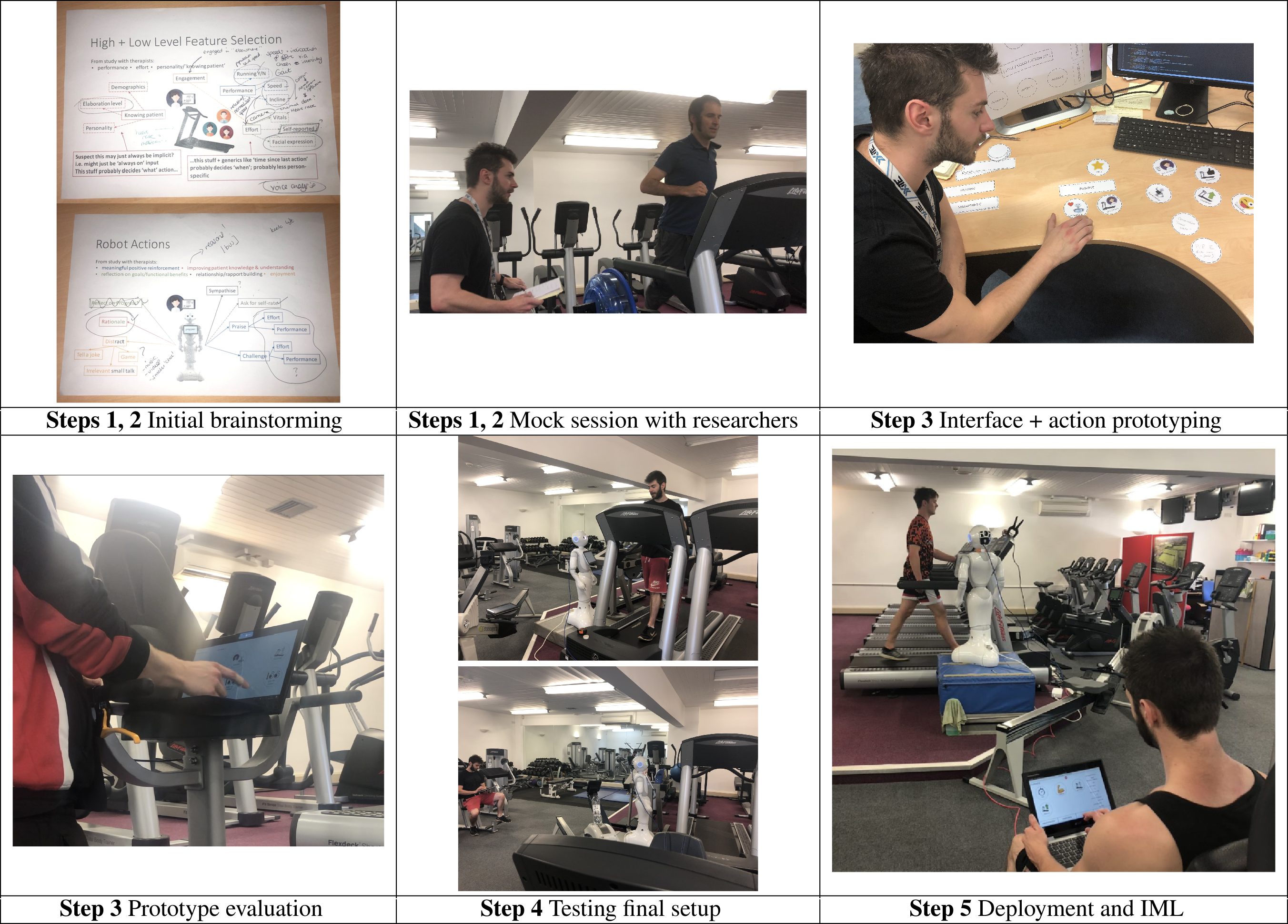}
        \end{center}
        \caption{Pictorial representations of the participatory design activities and final teaching setup undertaken in application of our method to~\cite{winkle2020insitu}'s robot fitness coach, as per Table~\ref{tab:studies} with reference to Steps 1-5 of our method as per Figure~\ref{fig:methodcomp}.}
        \label{tab:storyboard}
    \end{figure}
\end{landscape}

\end{document}